\documentclass[conference]{IEEEtran}
\IEEEoverridecommandlockouts

\usepackage{booktabs}
\usepackage{multirow}
\usepackage{graphicx}
\usepackage{float} 
\usepackage{xcolor} 
\usepackage{caption}

\usepackage{cite}
\usepackage{amsmath,amssymb,amsfonts}
\usepackage{algorithmic}
\usepackage{graphicx}
\usepackage{textcomp}
\usepackage{xcolor}
\def\BibTeX{{\rm B\kern-.05em{\sc i\kern-.025em b}\kern-.08em
    T\kern-.1667em\lower.7ex\hbox{E}\kern-.125emX}}
\begin{document}

\title{Short-form Text Rewriting with Phi Silica
\thanks{This work has been accepted for publication at IEEE ICAD. Copyright may be transferred without notice, after which this version may no longer be accessible.}

}
\author{\IEEEauthorblockN{Divya Tadimeti}
\IEEEauthorblockA{
\textit{Microsoft}\\
Cambridge, USA \\
dtadimeti@microsoft.com}
\and
\IEEEauthorblockN{Shawn Pan}
\IEEEauthorblockA{
\textit{Microsoft}\\
Cambridge, USA \\
shawnpan@microsoft.com}
\and
\IEEEauthorblockN{Sameera Lanka}
\IEEEauthorblockA{
\textit{Microsoft}\\
Redmond, USA \\
sameera.lanka@microsoft.com}
\and
\IEEEauthorblockN{Chenghui Zhou}
\IEEEauthorblockA{
\textit{Microsoft}\\
Cambridge, USA \\
zhouchenghui@microsoft.com}
\and
\IEEEauthorblockN{Sadid Hasan}
\IEEEauthorblockA{
\textit{Microsoft}\\
Cambridge, USA \\
	sadidhasan@microsoft.com}
}

\maketitle

\begin{abstract}
Short-form text rewriting is a constrained variant of paraphrasing in which limited context and high semantic density leave little room for variation. While large language models perform well on general paraphrasing, small language models (SLMs) often struggle with semantic fidelity and hallucination robustness in short-form settings. In this work, we present an empirical study of adapting an SLM, Phi Silica, for short-form rewrite through dataset curation, prompt distillation, parameter-efficient fine-tuning, and evaluation. We curate a dataset of short presentation-style text from public slide decks and use GPT-5-Chat both to generate rewrite supervision and to conduct LLM-as-a-judge evaluation. Our results show that finetuning improves semantic fidelity, reduces hallucinations, and increases preference win rate against GPT-5-Chat rewrites. The findings suggest that targeted adaptation for SLMs can substantially narrow the gap to cloud models and provide practical guidance for adapting SLMs to precision-critical rewrite tasks.

\end{abstract}

\begin{IEEEkeywords}
small language models, paraphrasing, rewrite, finetuning
\end{IEEEkeywords}

\section{Introduction}

Short-form text rewriting -- a constrained variant of paraphrasing -- presents unique challenges for language models due to limited context, dense semantics, and strict tolerance for meaning drift. Unlike long-form paraphrasing, minor lexical substitutions or syntactic changes in short text can materially alter intent, making semantic fidelity and natural flow especially important.

Large language models have demonstrated strong performance on paraphrasing and rewriting tasks, but their size and cost motivate interest in smaller models for constrained scenarios. However, small language models (SLMs) typically underperform on short-form rewrite without targeted adaptation, often exhibiting semantic drift, literal phrasing, or unnatural flow.

This paper presents a focused study of adapting an SLM, Phi Silica, for short-form text rewriting, using GPT-5-Chat as a high-quality reference model for supervision and evaluation.

Our contributions are threefold. First, we curate a large-scale dataset of short-form presentation text paired with high-quality rewrite supervision, designed to reflect realistic input characteristics. Second, we introduce and evaluate an LLM-as-judge framework - using GPT-5-Chat - to assess rewrite quality across multiple criteria and through pairwise comparisons. Third, we demonstrate that prompt distillation combined with LoRA fine-tuning significantly improves semantic fidelity, fluency, and resistance to hallucination over a baseline SLM, and provide analysis that informs future dataset curation and evaluation choices.
\section{Related Work}
\subsection{Small Language Models for Efficient Generation}
Recent advances in small language models (SLMs) have significantly narrowed the performance gap with large language models (LLMs) while maintaining lower inference costs. As a result, there has been a growing interest in identifying tasks where SLMs can serve as efficient alternatives to LLMs. 

Prior work has shown that, with task‑specific supervision, compact models can outperform zero‑shot prompting of large generative models, particularly in constrained or domain‑specific settings. Bucher and Martini (2024) demonstrate that fine‑tuned smaller models surpass zero‑shot LLMs on downstream NLP tasks, highlighting the limitations of prompt‑only adaptation\cite{bucher2024}. Gondara et al. (2025) further shows that while zero‑shot LLM performance degrades when task distributions diverge from pretraining data, fine‑tuned compact models maintain more stable performance, motivating supervised adaptation for efficiency‑critical tasks \cite{gondara2025}.

\subsection{On-Device and Short-Form Text Rewriting}
Recent work has begun to explicitly study text rewriting under device- and context-constrained settings. Zhu et~al.\ (2023) investigate on‑device text rewriting for short user messages, highlighting challenges arising from limited compute, restricted context, compact model capacity, and strict semantic requirements\cite{zhu2023}. The authors propose instruction-tuning and distillation strategies to adapt small models for rewriting tasks, and introduce \emph{MessageRewriteEval}, a human-labeled benchmark targeting short-form message rewriting scenarios. Their results demonstrate that task-specific adaptation substantially improves semantic fidelity and fluency for compact models, and that rewriting quality in short-text settings differs from outcomes observed on general paraphrasing benchmarks. This work provides direct evidence that short-form rewriting constitutes a distinct and challenging setting for efficient language model deployment.

More recent work has further explored the practical deployment of small language models under real-world constraints. Pham et al.\ (2024) introduce \emph{SlimLM}, a family of SLMs optimized for on-device document assistance tasks such as summarization and suggestion generation, and systematically study trade-offs between model size, context length, and inference latency on mobile hardware \cite{pham2024}. Their findings demonstrate that, when paired with task-specific fine-tuning, compact models can deliver competitive performance while enabling low-latency, privacy-preserving inference. While SlimLM focuses on general document assistance, our work isolates short-form rewriting as a precision-critical generation task where even minor semantic deviations can have outsized impact, highlighting a complementary but distinct challenge for efficient language models.

\subsection{Evaluation of Paraphrase Quality}
Evaluating paraphrasing quality, particularly under constrained rewriting conditions, remains a challenge. Standard reference-based metrics such as BLEU and ROUGE primarily capture lexical overlap and correlate poorly with human judgments of semantic equivalence and fluency. Shen et al. (2022) address this limitation with Parascore, a composite metric that measures semantic fidelity and lexical diversity \cite{shen2022}. Semantic‑aware evaluation better aligns with human preferences, especially for paraphrasing tasks requiring strict content preservation. These observations motivate evaluation protocols that move beyond surface‑level similarity for short‑form rewriting tasks.

Recent work has also examined the use of large language models as automatic evaluators for generative tasks. Gu et al.\ (2024) present a comprehensive review of the \emph{LLM-as-a-judge} paradigm, analyzing its methodology, strengths, and failure modes in text generation tasks \cite{gu2025}. Their study shows that rubric-guided LLM judges can correlate well with human judgments, but also identifies sensitivity to prompt design, output length, and near-tie cases -- issues that are especially pronounced for short-form generation tasks.

These findings align with our empirical observations that short-form rewrite evaluation exhibits variance when using small evaluation sets or loosely specified criteria. By adopting a fixed, length-stratified evaluation set and an explicit rubric for LLM-based judging, our work follows emerging best practices to mitigate instability in reference-free evaluation of constrained rewriting tasks.

\subsection{Parameter-Efficient Fine-tuning}
Several studies have examined parameter-efficient adaptation strategies for paraphrase generation. Jayawardena and Yapa (2024) introduce a sequence‑level knowledge distillation framework for paraphrasing, demonstrating that compact student models can closely match LLM teacher performance despite orders‑of‑magnitude reductions in model size \cite{jayawardena2024}. Techniques such as low‑rank adaptation (LoRA) further enable efficient fine‑tuning with minimal parameter updates, making small language models attractive candidates for constrained rewriting tasks where semantic precision, fluency, and robustness to hallucination are essential.

\section{Methodology}
We consider the task of short-form text rewriting. Given an input text segment, the model is expected to produce a rewritten version that improves clarity and fluency while preserving the original semantic intent. The rewrite must avoid hallucinations, excessive rephrasing, or unintended shifts in meaning. Because acceptable rewrites for short inputs are highly constrained, successful generation requires consistent adherence to the input meaning rather than stylistic creativity.

Our methodology therefore emphasizes controllability and measurement: training data are curated to constrain stylistic variation, prompts are distilled to reduce instruction sensitivity, and evaluation protocols are designed to surface small but systematic differences in rewrite quality. This framing allows us to assess whether improvements arise from learned rewrite behavior rather than prompt artifacts or evaluation variance.

\subsection{Dataset}\label{AA}

We curate a dataset of short presentation-style text sampled from publicly available slide decks. Textboxes are extracted and filtered to remove empty or near-empty spans. Inputs span a range of lengths, reflecting realistic presentation usage: very short (less than 40 characters, ~6\%), short (40–100 characters, ~23\%), medium (100–400 characters, ~55\%), long (400–600 characters, ~12\%), and very long (600–1000 characters, ~3\%). Each input is paired with a high-quality rewrite generated by GPT-5-Chat using deterministic decoding (temperature 0), yielding a conservative reference style that prioritizes semantic preservation. The final finetuning dataset is comprised of 93k rewrite pairs used as supervision for fine-tuning Phi Silica. 

For evaluation, we construct a standardized held-out set of approximately 1,000 textboxes. This set is sampled independently from training data and stratified by input length to reflect realistic usage scenarios and stratified by textbox length. All quantitative results in this work are reported using this fixed evaluation set.

\subsection{Models}

We study Phi Silica \cite{windowsblog_phisilica_2024} as the small language model under adaptation. Phi Silica is a compact language model (3.3B parameters) designed for efficient inference under constrained compute and memory budgets. Compared to large cloud-hosted language models, Phi Silica operates with substantially fewer parameters and reduced contextual capacity, which amplifies sensitivity to prompt design and data quality in precision-critical tasks such as short-form rewriting.

GPT-5-Chat is used as a high-quality cloud reference to generate rewrite supervision and to evaluate model outputs via LLM-as-judge scoring and pairwise preference comparison. GPT-5-Chat is not treated as a deployable baseline and is used solely as an upper-bound reference for rewrite quality.

We evaluate three variants of Phi Silica: the baseline pretrained model, a LoRA-finetuned full-precision (FP16) model, and a LoRA-finetuned quantized model.

\subsection{Prompt Design and Distillation}
Initial experiments used long, instruction-heavy rewrite prompts adapted from large-language-model workflows, including detailed task descriptions and many few-shot examples for paraphrasing. While effective for frontier models, this approach significantly degraded Phi Silica’s performance by overloading the limited context capacity of the small language model (SLM), introducing unnecessary instructional and example overhead during both data generation and inference.

To address this, we transitioned to a shorter, task-focused rewrite prompt for data generation that emphasizes semantic preservation and concise rewriting while removing excessive role conditioning and excess in-context examples. This prompt serves as the basis for constructing high-quality paraphrase pairs used during fine-tuning.

Following fine-tuning, we further distill the prompt used at inference time. By training the model on data generated with a richer rewrite specification, the model internalizes the desired rewrite behavior, allowing inference to use a minimal prompt with reduced runtime overhead.

\subsection{Finetuning Procedure}
We fine-tune Phi Silica using LoRA adapters applied to a dequantized base checkpoint. Training is performed using a batch size of 4, learning rate $5\times10^{-5}$, and LoRA rank 32 with dropout 0.3. Early stopping with a patience of 10 is applied, and models are trained for two epochs. In preliminary experiments, additional epochs yielded diminishing returns and, in some cases, regression in semantic fidelity, suggesting that short-form rewrite behavior can be learned efficiently with limited supervised adaptation.

During early experimentation, we observed training loss spikes and gradient explosion as a result of excessively long training sequences. To mitigate this issue, training samples are capped at 1000 characters.

\subsection{Quantization Considerations}
To assess robustness under reduced numerical precision, we evaluate both full‑precision and quantized versions of the fine‑tuned model. Fine‑tuning is performed in full precision, after which the learned weights are quantized without additional training. This analysis tests whether fine‑tuning gains persist after quantization; consistent performance suggests the learned improvements are stable and transferable to resource‑constrained downstream settings.

\section{Experimental Results and Discussion}
\subsection{Evaluation Protocol}
We evaluate rewrite quality using GPT‑5‑Chat as an LLM‑as‑a‑judge. We employ two complementary evaluation protocols: individual scoring and pairwise comparison. In the individual setting, the judge assigns dimension-specific scores - covering semantic similarity, hallucination, tone adherence, novelty and diversity, and grammar and fluency - to each rewritten output. These fine‑grained signals provide interpretable, absolute assessments that support controlled ablations and consistent comparisons across experimental conditions.
In parallel, we use pairwise preference judgments, where the judge compares two candidate rewrites and expresses a relative preference, capturing nuanced qualitative differences in a comparative form. Together, these two evaluation views offer mutually reinforcing perspectives on model behavior, combining calibrated absolute metrics with robust relative comparisons.

The evaluation criteria were informed by an initial structured human evaluation study. Human raters judged candidate rewrites along linguistically grounded dimensions including information preservation, naturalness, and conciseness. Analysis of this study revealed consistent perceptual boundaries: raters generally tolerated syntactic reordering and lexical substitution, but strongly penalized subtle shifts in factual content, changes in implied intent, or the introduction of unsupported inferences. These boundaries -- what humans intuitively consider acceptable variation versus meaning distortion -- directly informed both our fine-tuning objectives and evaluation design.

This human-in-the-loop study served as the foundation for calibrating our LLM-as-a-judge pipeline. We iteratively aligned the LLM’s scoring rubric and decision criteria to mirror human judgment patterns observed on a held-out set of manually annotated rewrites. This process ensured that the LLM evaluator did not overemphasize surface-level fluency or stylistic novelty, but instead reflected human-centered priorities around semantic fidelity, hallucination avoidance, and tone preservation.

For consistency across evaluations, the LLM-as-a-judge is provided with a structured rubric specifying the evaluation dimensions and scoring guidelines. Each rewrite is evaluated in isolation, without access to model identity or generation context, and scored using a fixed Likert-scale scheme. For pairwise comparisons, the evaluator selects the superior rewrite or indicates a tie when differences are negligible.

\subsection{Quantitative Results}
Fine-tuning substantially improves semantic fidelity, tone consistency, and hallucination robustness relative to the baseline Phi Silica model. Grammar and fluency are near ceiling for all models.

\subsubsection{Analysis of Individual Metric Evaluation}
The fine-tuned Phi Silica model demonstrates stronger semantic alignment, cleaner phrasing, and fewer hallucinations. Quantitatively, semantic similarity increases from 4.75 to 4.92, hallucination avoidance improves from 4.27 to 4.76, and tone consistency increases from 4.10 to 4.52, as shown in Table \ref{tab:metrics}. Grammar and fluency remain near ceiling across all models, indicating that fine-tuning primarily benefits meaning-preserving dimensions rather than surface correctness.

Novelty and diversity decrease modestly after fine-tuning (3.59 to 2.92). This change is consistent with the use of rewrite targets generated by GPT-5-Chat at temperature 0, which biases outputs toward conservative and deterministic phrasing. In this context, the reduction in novelty reflects successful imitation of the reference style rather than degradation in rewrite quality.

The quantized fine-tuned model closely matches the full-precision fine-tuned model across all metrics, suggesting that quantization does not meaningfully degrade rewrite quality for this task.

\subsubsection{Analysis of Pairwise Ranking Evaluation}

Pairwise ranking evaluation provides a complementary view of relative rewrite quality. As shown in Table~\ref{tab:ranking}, the baseline Phi Silica model wins only 22.34\% of comparisons against GPT-5-Chat on the final evaluation set. In addition, baseline win rates were observed to fluctuate depending on the sampled evaluation subset, often trending lower in the 12--13\% range. This variability highlights both the weakness of the baseline model and the sensitivity of short-form rewrite evaluation.

After fine-tuning on approximately 90k examples, the win rate increases to 29.66\%, with the quantized fine-tuned model achieving a similar win rate of 29.95\%. Importantly, fine-tuning also increases the tie rate relative to the baseline (5.95\% vs.\ 1.46\%), indicating a growing fraction of cases in which the model output is judged indistinguishable from GPT-5-Chat.

When considering both wins and ties, the fine-tuned Phi Silica model matches or exceeds GPT-5-Chat in 35.61\% of cases -- an increase of more than 11 percentage points over the baseline. This combined measure suggests that fine-tuning substantially narrows the quality gap between the SLM and the cloud reference, particularly for meaning preservation and grammatical fluency. Crucially, this improvement is maintained after quantization, supporting the robustness of the fine-tuned model.

\begin{table*}[t]
\centering
\caption{Pairwise preference ranking against GPT‑5‑Chat on the standardized evaluation set.}
\begin{tabular}{lccc}
\toprule
Model (vs. GPT‑5‑Chat) & Win (\%) & Loss (\%) & Tie (\%) \\
\midrule
Phi Silica (Baseline) & 22.34 & 76.20 & 1.46 \\
Phi Silica (Finetuned) & 29.66 & 64.29 & 5.95 \\
Phi Silica (Finetuned + Quantized) & 29.95 & 65.66 & 4.29 \\
\bottomrule
\end{tabular}
\label{tab:ranking}
\end{table*}

\begin{table*}[t]
\centering
\small
\caption{LLM-as-judge rubric scores (1--5) and total score (1--25) on the standardized evaluation set. Higher is better for all criteria.}
\begin{tabular}{lcccccc}
\toprule
\multirow{2}{*}{\textbf{Model}} &
\multicolumn{5}{c}{\textbf{Score (1--5)}} &
\multirow{2}{*}{\textbf{Total (1--25)}} \\
\cmidrule(lr){2-6}
& \textbf{Semantic} & \textbf{Hallucination} & \textbf{Tone} & \textbf{Novelty} & \textbf{Fluency} & \\
\midrule
Phi Silica (Baseline) & 4.75 & 4.27 & 4.10 & 3.59 & 4.95 & 21.66 \\
Phi Silica (Finetuned) & 4.92 & 4.76 & 4.52 & 2.92 & 4.97 & 22.09 \\
Phi Silica (Finetuned + Quantized) & 4.92 & 4.74 & 4.53 & 2.86 & 4.97 & 22.02 \\
GPT-5-Chat & 4.94 & 4.79 & 4.49 & 3.13 & 4.98 & 22.33 \\
\bottomrule
\end{tabular}
\label{tab:metrics}
\end{table*}

\subsection{Qualitative Analysis: Baseline SLM vs.\ GPT-5-Chat}
We first examine representative rewrite examples comparing the baseline Phi Silica model to GPT-5-Chat. These examples illustrate common failure modes of the baseline model prior to fine-tuning, particularly around semantic fidelity and natural flow.
\begin{table}[H]
\centering
\small
\begin{tabular}{p{0.95\linewidth}}
\toprule
\textbf{Example 1: Semantic Fidelity} \\
\midrule
\textbf{Original:} Look at your LSP’s track record on obligations. \\
\textbf{Phi Silica (Baseline):} Take a look at your LSP’s history with debts. \\
\textbf{GPT-5-Chat:} Review your LSP’s history of meeting its commitments. \\
\midrule
\textit{Observation:} The baseline model narrows the meaning of ``obligations'' to ``debts,'' introducing an unintended financial interpretation. GPT-5-Chat preserves the broader semantic intent. \\
\bottomrule
\end{tabular}
\end{table}

\begin{table}[H]
\centering
\small
\begin{tabular}{p{0.95\linewidth}}
\toprule
\textbf{Example 2: Clarity and Natural Flow} \\
\midrule
\textbf{Original:} What were your favourite sounds that you used or created? \\
\textbf{Phi Silica (Baseline):} What were the sounds you most enjoyed using or producing? \\
\textbf{GPT-5-Chat:} Which sounds did you enjoy using or creating the most? \\
\midrule
\textit{Observation:} The baseline rewrite is grammatically correct but reads as formal and less natural. GPT-5-Chat produces a more fluent and conversational phrasing. \\
\bottomrule
\end{tabular}
\end{table}

\noindent\textbf{Summary.}
Across these examples, the baseline Phi Silica model frequently alters meaning through overly literal substitutions or produces grammatically correct but unnatural phrasing. These issues motivate targeted fine-tuning.

\subsection{Qualitative Analysis: Effect of Fine-Tuning}

We next compare baseline and fine-tuned Phi Silica outputs to assess how fine-tuning affects rewrite quality relative to GPT-5-Chat.

\begin{table}[H]
\centering
\small
\begin{tabular}{p{0.95\linewidth}}
\toprule
\textbf{Example 3: Post-Fine-Tuning Comparison} \\
\midrule
\textbf{Original:} The strength of our whole A.A. service structure starts with the group and with the general service representative (G.S.R.) the group elects. I cannot emphasize too strongly the G.S.R.’s importance. \\
\textbf{Phi Silica (Baseline):} Our entire Al-Anon program’s robustness is anchored by the group and elected general service representative (G.S.R.). I cannot stress enough the critical role of the G.S.R. enough. \\
\textbf{Phi Silica (Finetuned):} The foundation of our entire A.A. service system begins with the group and the general service representative (G.S.R.) chosen by the group. I cannot stress enough how vital the G.S.R. is to the overall structure. \\
\midrule
\textit{Observation:}The baseline model introduces entity‑level hallucination by expanding “A.A.” to “Al‑Anon” and produces redundant phrasing that degrades fluency. Fine‑tuning resolves both issues, yielding a rewrite that preserves the original entity and more closely matches the semantic fidelity and tone of GPT‑5‑Chat. \\
\bottomrule
\end{tabular}
\end{table}

\begin{table}[H]
\centering
\small
\begin{tabular}{p{0.95\linewidth}}
\toprule
\textbf{Example 4: Vocabulary and Style} \\
\midrule
\textbf{Original:} Interchangeable, Not Interchangeable, New for Old, Old for New \\
\textbf{Phi Silica (Baseline):} Swappable, Not Swappable, New Old, Old for New \\
\textbf{Phi Silica (Finetuned):} Exchangeable, Incompatible, Replacement, Upgrade \\
\midrule
\textit{Observation:} The baseline output largely mirrors surface structure, while the fine-tuned model introduces more expressive and context-appropriate vocabulary. \\
\bottomrule
\end{tabular}
\end{table}
\noindent\textbf{Summary.}
During development, we observed that short-form rewrite evaluation is highly sensitive to both training data composition and evaluation set construction. Early experiments using smaller training sets (under 50k examples) and limited evaluation samples often suggested larger fine-tuning gains, but these results varied substantially across different evaluation draws.

Further investigation revealed that small evaluation sets amplified variance in pairwise preference outcomes, particularly when comparing near-equivalent rewrites differing only in phrasing.

After adopting a fixed, length-stratified evaluation set of approximately 1,000 examples, fine-tuning gains became smaller in magnitude but more consistent. We therefore report only results derived from this stabilized configuration. While this conservative reporting likely understates peak achievable gains, it provides a more reliable estimate of real-world performance improvements.

These observations highlight that, for short-form rewriting, evaluation stability is as critical as model architecture or fine-tuning strategy. Apparent improvements can be driven by dataset artifacts unless controlled through careful sampling and validation.

Based on qualitative observations during development, rewrite behavior appears to
vary with input length. Very short textboxes are often susceptible to literal
substitutions that narrow meaning, while medium-length inputs frequently benefit
from fine-tuning due to sufficient contextual cues without overwhelming model
capacity. Longer textboxes, which are less common in practice, remain challenging
for SLMs due to increased compositional complexity and weaker supervision density.

These findings indicate that future improvements may benefit more from targeted dataset curation -- such as emphasizing failure-prone short inputs or underrepresented semantic patterns -- than from further indiscriminate dataset scaling.

\subsection{Implications for On-Device and Constrained Generation}
These results have implications beyond the specific model studied. Short-form rewriting is representative of a broader class of precision-critical generation tasks -- including message refinement, text simplification for readability, and UI-constrained text assistance -- where outputs must remain faithful under limited context. Our findings suggest that, in such settings, careful prompt distillation and targeted data curation may yield larger practical gains than further model scaling. Moreover, the preservation of quality after quantization indicates that parameter-efficient adaptation can be compatible with real-world deployment constraints, supporting the feasibility of high-quality on-device language assistance.

\section{Conclusion}
We present a learning-focused study of adapting a small language model, Phi Silica, for short-form text rewriting. Through dataset curation, prompt distillation, and parameter-efficient fine-tuning, we significantly improve semantic fidelity and natural flow relative to the baseline model. Our analysis reveals diminishing returns from dataset scaling and underscores the importance of targeted data composition, particularly with respect to input length distribution. These findings offer practical guidance for future efforts to adapt SLMs to precision-critical rewriting tasks. The results suggest that well-targeted adaptation can enable practical, high-quality rewriting in on-device and other constrained inference scenarios, where latency, privacy, and reliability are paramount.
\

\section*{Acknowledgment}
We thank Sunando Sengupta, Dimitrios Mallios, Marat Saidov, Teo Sarkic, Ivan Razumenic, Milomir Stefanovic, Henry Jackson-Flux, Milos Stojanovic, and others for providing guidance and access to shared resources and tooling that enabled this effort. We also thank Vishal Chowdhary, Michael Bentley, and Joshua Burkholder for helpful discussions and feedback. Finally, we thank Sushmitha Muppa and Ziran Min for engineering support during the course of this work.

\bibliographystyle{IEEEtran}
\bibliography{references} 

\end{document}